\documentclass{article}

\PassOptionsToPackage{numbers,square,sort&compress}{natbib}
\usepackage[final]{neurips_2021}

\usepackage[utf8]{inputenc} 
\usepackage[T1]{fontenc}    
\PassOptionsToPackage{hyphens}{url}
\usepackage{hyperref}       
\usepackage{booktabs}       
\usepackage{nicefrac}       
\usepackage{microtype}      
\usepackage{xcolor}         

\usepackage{amsmath,amsfonts,amsthm,amssymb}
\usepackage{bm,cases}
\usepackage{float,caption}    
\usepackage{graphicx,wrapfig} 
\usepackage{enumitem}         
\setenumerate[1]{leftmargin=2em, itemsep=0pt, partopsep=0pt, parsep=\parskip, topsep=0pt}
\setenumerate[2]{leftmargin=2em, itemsep=0pt, partopsep=0pt, parsep=\parskip, topsep=0pt}
\setitemize[1]{leftmargin=2em, itemsep=0pt, partopsep=0pt, parsep=\parskip, topsep=0pt}
\setitemize[2]{leftmargin=2em, itemsep=0pt, partopsep=0pt, parsep=\parskip, topsep=0pt}

\title{Calculus of Consent via MARL: Legitimating the Collaborative Governance Supplying Public Goods}

\author{%
  Yang Hu \\
  Tsinghua University \\
  Beijing, China \\
  \texttt{huy18@mails.tsinghua.edu.cn} \\
  \And
  Zhui Zhu \\
  Tsinghua University \\
  Beijing, China \\
  \texttt{zhuz18@mails.tsinghua.edu.cn} \\
  \And
  Sirui Song \\
  McGill University \\
  Montréal, Canada \\
  \texttt{sirui.song@mail.mcgill.ca} \\
  \And
  Xue Liu \\
  McGill University \\
  Montréal, Canada \\
  \texttt{xueliu@cs.mcgill.ca} \\
  \And
  Yang Yu \\
  Tsinghua University  \\
  Beijing, China \\
  \texttt{yangyu1@tsinghua.edu.cn} \\
}

\begin{document}

\maketitle

\begin{abstract}
  Public policies that supply public goods, especially those involve collaboration by limiting individual liberty, always give rise to controversies over governance legitimacy. Multi-Agent Reinforcement Learning (MARL) methods are appropriate for supporting the legitimacy of the public policies that supply public goods at the cost of individual interests. Among these policies, the inter-regional collaborative pandemic control is a prominent example, which has become much more important for an increasingly inter-connected world facing a global pandemic like COVID-19. Different patterns of collaborative strategies have been observed among different systems of regions, yet it lacks an analytical process to reason for the legitimacy of those strategies.
In this paper, we use the inter-regional collaboration for pandemic control as an example to demonstrate the necessity of MARL in reasoning, and thereby legitimizing policies enforcing such inter-regional collaboration. 
Experimental results in an exemplary environment show that our MARL approach is able to demonstrate the effectiveness and necessity of restrictions on individual liberty for collaborative supply of public goods. Different optimal policies are learned by our MARL agents under different collaboration levels, which change in an interpretable pattern of collaboration that helps to balance the losses suffered by regions of different types, and consequently promotes the overall welfare. Meanwhile, policies learned with higher collaboration levels yield higher global rewards, which illustrates the benefit of, and thus provides a novel justification for the legitimacy of, promoting inter-regional collaboration.
Therefore, our method shows the capability of MARL in computationally modeling and supporting the theory of calculus of consent, developed by Nobel Prize winner J. M. Buchanan.



\end{abstract}

\section{Introduction}
As brilliantly revealed in J. M. Buchanan's celebrated book \textit{The Calculus of Consent} \cite{buchananCOC},  collective actions are always formed by (probably conflicting) individual actions, during which process some individuals should rationally alienate or waive their rights of liberty based on the calculus of consent. This outstanding theory of public choice applies to the supply of public goods, which generally requires the collaboration among multiple parties that could restrict the liberty of some parties for the sake of public interest. Thus, the controversy between the extent of liberty restriction and the necessary level of collaboration is naturally raised for debate, and governors will face the problem of legitimacy whenever they fail to explain the necessity and intensity of collaborative policies.

It should be noticed that the making of optimal policies and their legitimacy explanations involve delicate quantified calculation of individual and public interests, or more specifically in the scope of public goods supply, the balance of costs and benefits. The calculation is of such high computational complexity that is beyond the capability of human instincts, which impairs the legitimacy and credibility of policy-makers, and thus restricts them from fully functioning to pursue the greatest good for the public. Fortunately, the rapidly developing computational tools for decision making, such as optimal control and reinforcement learning, have enabled us to find optimal solutions for complicated problems, so they can effectively serve as convincing justifications for policy-makers.

There is no doubt that the most recent large-scale debate over the supply of public goods is about COVID-19 pandemic control policies. In the highly globalized and interconnected world today, pandemic outbreaks have proven to be more devastating to human society than any time in history --- they not only cause millions of deaths around the world, but also jeopardize the global value chain (GVC) \cite{fukunari2020GVC} and give rise to billions of dollars of economic losses. Economists have realized that the control of infectious diseases falls into the category of global public goods (GPGs) that cannot be efficiently supplied without intensive inter-regional collaboration \cite{brown2020GPG}. And here naturally comes the problem of legitimacy: how can a central government convince its local governments, or an international association convince its member countries, that enforcing strict lockdown policies is an appropriate and necessary measure to take? Why should the latter be willing to carry out those policies that might harm their own economic interests?

Indeed, regions are not equally willing to collaborate at the cost of their own benefits --- regions in a collaborative system are expected to enforce strict lockdown policies to guarantee global well-being, even if they will damage the local economy; regions in a self-centered system, however, will more likely carry out loose restrictions to save the local economy, at the cost of wider-spread diseases and global utility losses which they do not really care.
As a result, different behavior patterns will be observed in systems of regions that vary greatly in the the willingness to collaborate, and thus a legitimate policy in one system might not be so in another --- strict lockdown policies can be easily enforced in unitary governments, but harder in federal governments; vaccines can be efficiently distributed within a country, but significantly less smoothly within an international association. Differences in the willingness to collaborate root in the difference in social capitals, which plays a decisive role in the organization of collaborative supply of public goods.

Such ``willingness'' should be emphasized in the analysis of inter-regional pandemic control policies. It can be quantitatively characterized by the proportion of global utility appearing in each region's policy-making objective, which we shall refer to as \textit{collaboration levels}. Differences in collaboration levels may root from different sources --- it could be the natural consequence of a coherent and unified political structure (e.g., as in China) that always acts in pursuit of greatest global utility; or it could be due to the political relationship and economic inter-dependence (e.g., as in the EU), so that the high externality of pandemic control prompts the regions to care more about the global utility.

Finding effective pandemic control measures involves both epidemiological and socioeconomic considerations. On the epidemiological side, research has revealed the effect of human mobility on the COVID-19 pandemic \cite{Wuhan, US}, pointing out the fact that respiratory infectious diseases like the COVID-19 can hardly be controlled without strict international restrictions on the mobility of population and commodities. However, in the socioeconomic perspective, strict lockdown policies will probably damage local economy \cite{FLD}, leading to unfavorable consequences like unemployment, supply shortage or even depression. Therefore, policy-makers face an inherent trade-off between pandemic control and economic well-being, and the equilibrium is determined by the \textit{pandemic-tolerance level} (whether it has abundant medical resources to endure a long-term pandemic outburst) and the \textit{lockdown-tolerance level} (whether its economic structure can survive a long-term lockdown).

Computational approaches are playing an increasingly important role in search of optimal pandemic control policies. For example, \citeauthor{Optimal} proposed an optimization-based decision-making framework to calculate optimal pandemic control policies \cite{Optimal}. Artificial intelligence tools such as Multi-Agent Reinforcement Learning (MARL) algorithms have also be employed to allow more complex models, balancing health and economic costs \cite{DURLECA, RLepidemic}. However, all of these studies focus on centralized control policies only, so that they do not take into consideration how multiple independent regions should and would collaborate to fight against the pandemic. Therefore, we feel motivated to invent new computational models that helps to analyze the underlying patterns of inter-regional collaboration, which also helps to find and legitimize optimal pandemic control policies.

The structure of the multi-region collaborative pandemic control problem bears several challenges, such as intensive coupling among regions brought by pandemic spread, the exponential explosion nature of pandemic spread, and the instability of observed environment due to inderdependency of regional policies.
To address these inherent challenges, we design an MARL model called Intelligent Region Collaboration (IRC), where each region is represented by a learning agent, and each agent's action is to block a proportion of traffic from any other regions into it, which abstracts the mobility restriction policy in reality. Regions are assigned different \textit{pandemic-tolerance levels} and \textit{lockdown-tolerance levels} to capture the major factors that influence policy-making. To reflect different collaboration levels, each agent is assumed to receive both an individual reward and a global reward, and the trade-off weight between these two terms, known as the \textit{reward mixing ratio}, quantifies collaboration levels. The MARL model is trained with a specifically designed actor-critic algorithm that ``decouples'' the highly unstable environment faced by each agent into a stable local environment learned by the agent itself, and a stable global environment learned globally.

The IRC method proposed in this paper serves as an effective tool to find optimal pandemic control policies under different inter-regional collaboration levels. These policies not only help policy-makers in the real world, but also provide a perspective to explore the influence of different tolerance levels and collaboration levels on agents' behavioral patterns and the final outcomes. For this purpose, we construct an exemplary environment with agents of different tolerance levels, and observe the policy learned by our model under different collaboration levels. The learned policies are evaluated by multiple metrics, compared with two heuristic baseline policies. Experimental results show that different collaboration patterns appear under different collaboration levels in an interpretable way. It is observed that, when a region only cares about its own interest, its optimal pandemic control policy is almost completely based on its own tolerance levels, and thus the amount of blocked traffic is highly imbalanced among different regional types; on the other hand, as global interests become dominant in a region's reward composition, it is more likely for a region to sacrifice a small portion of its local reward by blocking extra amount of traffic into it, in exchange for lower lockdown penalties of more vulnerable regions. The collaboration is also characterized by more evenly distributed mobility among different types of regions.

The contribution of this paper is that we design an MARL model to find optimal region-collaboration pandemic control policies, which contains an adjustable parameter that quantifies the collaboration level among regions, or more specifically, the proportion of global reward in each agent's mixed reward. In this way, we obtain a new perspective to analyze the collaboration behavior of regions with different types under different collaboration levels, and by examining the patterns in the learned optimal policies we can better understand the underlying logic of inter-regional collaboration, and thus provide new justifications for the legitimacy of collaborative governance supplying public goods. 

\section{Related Works}
\vspace*{-5pt}
In this section, we discuss the research that is the most pertinent to the work in this paper.

\vspace*{-5pt}
\subsection{Multi-Agent Reinforcement Learning}
Reinforcement learning (RL) has long been an active research field for many decision-making problems \cite{RL, RL2}. Despite its success in single-agent settings, Multi-Agent Reinforcement Learning (MARL) that deals with multiple autonomous agents is much more complicated, and has seen a rapid development in the past decade \cite{MARL2, MARL3}. State-of-the-art MARL algorithms are able to show human-level performance in complex multi-agent games such as DOTA \cite{OpenAI_Dota} and StarCraft II \cite{Vinyal2017Starcraft}.

In the typical cooperative MARL setting, the environment responds to the aggregate action of all agents with a single global reward, and agents take joint efforts to maximize total global reward. Therefore, a natural idea is to learn the decomposition of the global reward first, and then apply value-based RL algorithms for each agent, which is adopted by popular algorithms like VDN \cite{VDN2018}, QMIX \cite{QMIX2018}, QTRAN \cite{QTRAN2019}, etc. Another approach is to adapt policy gradient methods for multi-agent settings, among which the most successful algorithms are MADDPG \cite{MADDPG}, COMA \cite{COMA} and MAAC \cite{MAAC}. There is also much research investigating the cooperative behaviour of MARL agents in various environments \cite{Tampuu2017MARL, Gupta2017MARL}.  

The multi-agent setting in this paper is different from the typical cooperative setting in most existing literature. Our pandemic control task is special in that agents are allowed to retrieve their own local rewards in response to their actions, and thus the value decomposition network is unnecessary. Meanwhile, collaboration level is abstracted as an adjustable hyperparameter, so that collaborative behavior can be controlled rather than only observed.

\subsection{Pandemic Control}
In epidemiological studies, the Susceptible-Infected-Recovered (SIR) model is a widely used mathematical model for pandemic transmission, on which many research on pandemic control strategies are based \cite{SIR}. It is also modified for more realistic modelling: SEIR model added an Exposed state to make the model more generalizable \cite{SEIR}, while DURLECA proposes the SIHR model to capture the challenges posed by asymptomatic infections \cite{DURLECA}.

Based on these pandemic transmission models, researchers have discussed how to find the optimal pandemic control strategy in different settings using traditional optimization approaches \cite{SIR2,SIR3,SEIR2}. More recently, the application of RL methods is proposed, where the actions of all regions are determined by one agent in a centralized manner \cite{DURLECA}. Compared with these existing studies, our IRC method is the first to apply MARL methods for optimal pandemic control policies, and to investigate the behaviour patterns of different regions under different collaboration levels.

\section{Modeling and Method}
\subsection{Environment Modeling}
The environment model focuses on the traffic (or \textit{mobility}) and pandemic transmission among different regions. Therefore, each region can be simply regarded as a single node within a fully connected graph, and all within-node details are excluded from our model.

\textbf{Mobility.} The amount of mobility demand $M_{\text{d},i,j}^t$ between each pair of regions $i,j$ at each time step is pre-determined. Each region regulates its pandemic control policy by determining the proportion of incoming mobility demand that can be fulfilled (note that traffic is controlled by destination region rather than the departure region). Therefore, region $j$ is allowed to determine $p_{i,j}^t \in [0,1]$ for all $i \neq j$, and fulfills only that proportion of the total mobility demand. A more rigorous formulation of mobility demand and actions can be found in Appendix \ref{appendix:mobility}.

It is reasonable to abstract pandemic control policy as incoming mobility restrictions. Mobility between regions reflects the quantity of economic activities, and mobility restrictions will lead to losses of economic interests; mobility of people and commodities can be easily controlled by region governments via border and public transportation controls; mobility is also closely related to pandemic control as unrecognized infections will spread the virus to the destination. Therefore, in our model, agents are expected to maximize the fulfillment of mobility demands, but also wisely restrict mobility to reduce the spread of pandemic simultaneously.

\textbf{Pandemic transmission.} The pandemic transmission model is based on the Susceptible-Infected-Hospitalized-Recovered (SIHR) model proposed in \cite{DURLECA}. The \textit{pandemic state} of region $i$ at time step $t$ is $E^t_i= (S^t_i, I_i^t, H_i^t, R_i^t)$, which consists of the number of susceptible, infected, hospitalized, and recovered population within that region. Agents are able to observe the number of hopitalization and recovery, but cannot distinguish between susceptible (healthy) people and infected people that are both asymptomatic. Therefore, the \textit{visible pandemic state} of region $i$ at time step $t$ is $E^t_{\text{v},i}= (S^t_i+I_i^t, H_i^t, R_i^t)$. The detailed model for pandemic transmission is deferred to Appendix \ref{appendix:transmission}.

\subsection{Problem Formulation}
The target of our multi-region collaborative pandemic control problem is two-fold --- agents are expected to minimize the spread of infection, and fulfill the maximum proportion of mobility demand at the same time. Therefore, the \textit{local reward} $R_i$ of region $i$ is determined by two factors: how much mobility demand into this region is fulfilled, and how serious the pandemic state is in this region. More rigorously, each region $i$ receive a local reward $R_i$ for its pandemic situation at each time step, which is a sum of the \textit{pandemic-spread cost} $C_{\text{p}, i}$ and the \textit{mobility-control cost} $C_{\text{m},i}$
\begin{equation}
  R_i=-C^t_{\text{p},i}-C^t_{\text{m},i},
\end{equation}
The \textit{global reward} $R_\text{g}$ (the common interest of all regions) is the sum of all local rewards
\begin{equation}
  R_\text{g} = \sum_{i=1}^{n} R_i = \sum_{i=1}^{n} \left( -C^t_{\text{p},i}-C^t_{\text{m},i} \right).
\end{equation}
Here costs $C^t_{\text{p}, i}$ and $C^t_{\text{m}, i}$ grow exponentially with regard to the number of hospitalized patients ($H^t_i$) and an accumulated amount of blocked mobility, respectively. Two characteristic hyperparameters, namely \textit{pandemic-tolerance level} $H_{0,i}$ and \textit{lockdown-tolerance level} $L_{0,i}$, represent the exponential grow rate mentioned above. The pair $(H_{0,i}, L_{0,i})$ is also referred to as \textit{region types} since they sufficiently reflect a region's resilience for pandemic outbreak and lockdown --- the higher the hyperparameter, the more resilient a region is.
Detailed definitions of these costs and tolerance levels are deferred to Appendix \ref{appendix:tolerance-levels}.

Although the two tolerance levels seem highly abstract and artificial, they are meaningful in practice. Countries and cities have different supply of medical resources, and those with more medical resources tend to be more resilient to a pandemic outbreak. Meanwhile, regions whose economy rely heavily on travellers and imported commodities will be more reluctant to carry out complete lockdown policy, as it may hurt the economy even more severely than the pandemic outbreak. Therefore, the simple abstraction does capture the main conflicting factors in pandemic control.

Now we can follow the multi-region collaborative pandemic control problem. The objective of each agent is a weighed sum of local and global rewards, namely
\begin{equation}
  p^{t}_i = \mathop{\arg\max}_{p} \big\lbrace R_i(p; M^t_{\text{d},i}, E^t_i)+\alpha R_\text{g}(p; M^t_{\text{d}},E^t) \big\rbrace.
\end{equation}
where we introduce a \textit{reward mixing ratio} $\alpha$ to represent the proportion that the global reward accounts for in the mixed reward $\tilde{R}_i = R_i + \alpha R_\text{g}$.
\subsection{Intelligent Region Collaboration (IRC)}

\begin{figure}[h]
  \centering
  \includegraphics[width=0.9\linewidth]{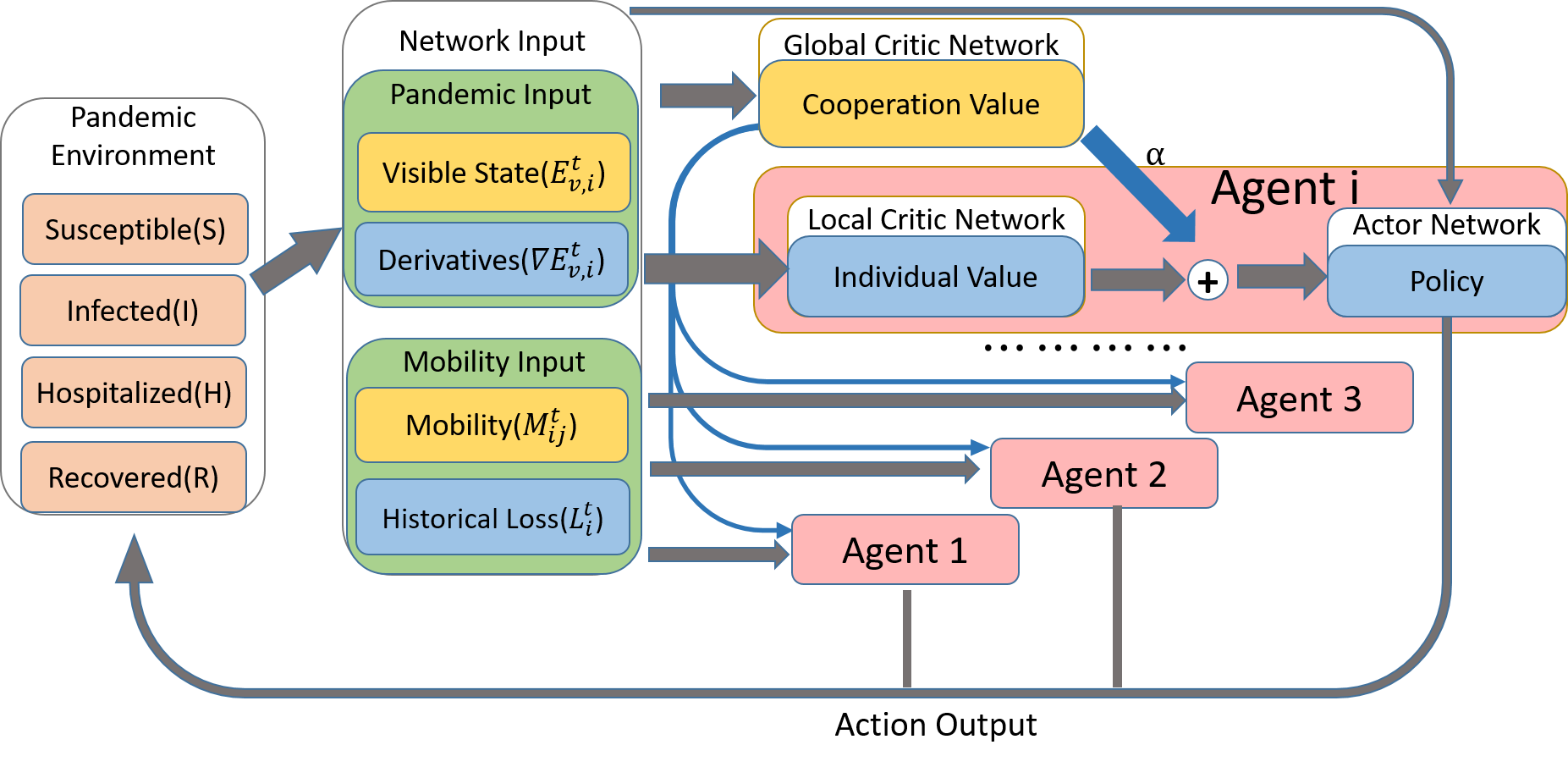}
  \caption{Schematic diagram of Intelligent Region Collaboration (IRC) model.}
\end{figure}


To solve the multi-region collaborative pandemic control problem, we design an MARL method called Intelligent Region Collaboration (IRC). The IRC method implements a multi-agent actor-critic RL algorithm enhanced by GNN networks. Each region is modeled as an RL agent. For agent $i$ at time step $t$, it first acquires an observation consisting of the visible pandemic state $E_{\text{v}}^t$ and mobility demands $(M^t_\text{d},M^{t+1}_\text{d},M^{t+2}_\text{d}, M^{t+3}_\text{d})$ for a period of 4 time steps from the environment; then based on this information, the agent determines its action $p^{t:(t+3)}_i$ for the next 4 time steps. Note that control actions are carried out every several time steps, which makes practical sense since changes in mobility demands and control policies can only be applied periodically in reality.

Details of our RL framework are presented below:
\begin{itemize}
    \item \textbf{State.} The state includes the visible pandemic state $E_{\text{v}}^t$ and its temporal derivative $\nabla_t E_{\text{v}}^t$.
    
    \item \textbf{Action.} The action of each RL agent $i$ is a column vector $p^t_{i} = (p^t_{1,i}, \cdots, p^t_{n,i})^{\top}$ ($n$ is the number of regions), where $p^t_{j,i}$ is the fulfilled proportion of mobility demand $M^t_{\text{d},j,i}$.

    \item \textbf{Reward.} Agent $i$ receives a \textit{mixed reward} $\tilde{R}_i = R_i + \alpha R_\text{g}$ as defined previously.
    
    \item \textbf{Learning Algorithm.} The IRC method is based on the GNN-enhanced Deep Deterministic Policy Gradient (DDPG) method used in DURLECA \cite{DURLECA}. The GNN is used to estimate the future pandemic status from the observations, which can utilize the underlying graph structure. Each agent is trained by the DDPG algorithm introduced in \cite{DDPG}.
\end{itemize}

To deal with the challenge of mutually coupled and volatile environment observed by each agent that obstructs efficient training, critics in our IRC model are specially designed to utilize physical decomposablity of local and global rewards in our setting. Therefore, instead of employing a single critic for each agent and introducing an extra value-decomposition network, as in standard MADDPG agents \cite{MADDPG}, we employ an individual \textit{local critic} for each agent and an additional shared \textit{global critic} for all agents to learn local and global rewards separately.
More specifically, each DDPG agent in our IRC method is equipped with an actor and a local critic, and has access to a shared global critic.
The local critic is customized for each agent, which learns the $Q$-function associated with the local reward $R_i$; the global critic is shared by all agents, which learns the $Q$-function associated with the global reward $R_\text{g}$. 
The actor generates actions that maximizes the weighed sum of the local critic output and the global critic output.
In this way, the highly unstable environment faced by each agent is decoupled into a stable local environment learned by the agent itself and a stable global environment learned by all the agents from a global view. 

\section{Experiment and Evaluations}\label{sec:experiment}
In this section, we present the experimental results in a representative setting, in order to reveal the patterns of collaboration behavior among regions with different types under different collaboration levels, based on the optimal policies learned by our IRC method. We first describe the setup and evaluation metrics used in the experiments, and then present the results and our interpretations.

\subsection{Experimental Settings}
The experiment focuses on the early ``outbreak'' stage of pandemic transmission, where most regions contain few infected people, but one specific \textit{source region} (labelled as region \#0) is suffering from an already out-of-control pandemic outbreak ($R_0 > 1$), and infections will spread from the source region to other regions. This stage is critical to successful pandemic control. Since we care more about preventing the vast pandemic spread among all the regions than controlling infections in the source region (which is configured to be impossible, as $R_0 > 1$ in region \#0), we specially exclude the pandemic penalty from region \#0's local reward.  

\textbf{Environment and region types.} With the main purpose of revealing the influence of tolerance levels and collaboration level on the outcome, the artificial setting contains only 5 regions. Each region has an initial population of 10,000,000 people, a daily traffic of 5,000 people for each travelling route, and 2,000 initial infections in source region \#0. Apart from the source region, other regions (\#1 through \#4) are designed to demonstrate different strategies learned by agents of different region types. Here we assume the tolerance levels are set to be either ``high'' ($H_+$ or $L_+$) or ``low'' ($H_-$ or $L_-$), where $H_+ > H_-$ and $L_+ > L_-$. It is evident that 4 non-source regions are sufficient to cover all possible region types, namely $(H_+, L_+), (H_+, L_-), (H_-, L_+), (H_-, L_-)$, respectively.

We point out that region types are abstract hyperparameters specific to our pandemic model, so that it is their relative relationship rather than absolute values that matter in our illustrative experiment, which justifies the two-level setting. For the sake of implementation, we assign the specific values as $H_+ = 0.003$, $H_- = 0.001$, $L_+ = 72$ and $L_- = 24$. The source region \#0 is special in that its pandemic situation is not concerned, so we assign to it a lockdown-tolerance level $L_{0,0} = 0.05$.

\textbf{Collaboration levels.} To investigate agents' behavioral patterns under different collaboration levels, our IRC models are trained with 3 different reward mixing ratios $\alpha \in \{0.01, 0.40, 10.0\}$. For each configuration, the training procedure is repeated for 10 times with different random seeds, and the model that yields the highest reward is selected for further evaluation and analysis.

\subsection{Evaluation Protocols}

\textbf{Baselines.} Our MARL models are compared against the following two baselines (or conceptually ``expert policies''), which are abstracted from common and most easily implementable practices.
\begin{itemize}
    \item \textit{Fixed policy.} Each region allows a fixed proportion $p_{\text{fix}}$ of all incoming traffic for entrance. The proportion is global in the sense that it works for all regions and throughout the pandemic period. This abstracts typical control strategies like permanently restricting international flights. 

    \item \textit{Threshold policy.} This is a more flexible strategy that dynamically ``respond to'' the changing pandemic situation, where lockdown is enforced when there are many observed (i.e., hospitalized) patients, indicating that the pandemic situation is worsening, and the cumulative mobility loss is still acceptable, indicating that the lockdown will not hurt economy too seriously. More specifically, the policy sets each action $p^t_{i,j}$ by the rule
    \begin{equation}
      p^t_{i,j} = \begin{cases}
        0 & H^t_i > H_{\text{th}} ~\text{and}~ L^t_i < L_{\text{th}} \\
        1 & \textrm{otherwise}
      \end{cases}.
    \end{equation}
\end{itemize}

\textbf{Metrics.} Several metrics reflecting the effectiveness of pandemic control and the extent of collaboration are used to quantitatively evaluate the performance of our models against the baselines:
\begin{itemize}
  \item \textit{Mean global reward $\overline{R}_{\text{g}}$.} This is a direct revelation of the performance that the model is trained to optimize. Here we only need the mean of global reward, since the local rewards of all agents sum to the global reward.
  
  \item \textit{The mean $\overline{H}$ and maximum $H_{\text{max}}$ number of total hospitalized patients.} These two metrics represent the pressure on local medical services brought by the pandemic control policy, which reflect the feasibility of the pandemic control policy in medical sense.
  
  \item \textit{Mean action $\overline{p}$}. Mean action is defined as the ratio of total actual mobility over total mobility demand into a region. It represents the strictness of inter-regional lockdown enforced by a region, and acts as an indicator for the economic loss induced by the pandemic control policy.
  
  \item \textit{Type-wise analysis.} To analyze the collaboration behavior of agents with different regional types, the mean hospitalization $\overline{H}$ and mean mobility $\overline{p}$ within regions of the same type are also calculated, and presented in the form of 2-by-2 matrices and/or radar plots.
\end{itemize}

\subsection{Experimental Results}

The results of baseline polices and our IRC models are shown in \autoref{table:toy-outbreak-table}. It is clear that, regardless of the mixing ratio $\alpha$, our IRC models outperform both baselines by receiving larger global rewards, higher mobility and lower mean hospitalization rate. Therefore, our IRC models achieve better performance at balancing the pandemic cost and lockdown cost of each region, and at balancing the demand of different regions.

  \begin{table}[htbp]
    \centering
    \caption{The performance of different models.}\label{table:toy-outbreak-table}
    \begin{tabular}{cc|cccc}
      \specialrule{1.5pt}{0pt}{0pt}
      \textbf{model} & \textbf{parameter} & \textbf{$\bm{\overline{R}_{\text{g}}}$} & \textbf{$\bm{\overline{H}}$} & \textbf{$\bm{H_{\textbf{max}}}$} & \textbf{$\bm{\overline{p}}$} \\
      \specialrule{0.4pt}{0pt}{0pt}
      Fixed & $p_{\text{fix}} = 0.5$ & $64.16$ & $88.29$ & $276.68$ & $0.5000$ \\
      Threshold & {\footnotesize $H_{\text{th}} = 1, L_{\text{th}} = 168$} & $88.82$ & $92.65$ & $202.17$ & $0.8198$ \\
      \specialrule{0.4pt}{0pt}{0pt}
      IRC & $\alpha = 0.01$ & $\bm{103.89}$ & $59.46$ & $373.25$ & $0.7753$ \\
      IRC & $\alpha = 0.40$ & $\bm{106.97}$ & $31.29$ & $327.90$ & $0.7090$ \\
      IRC & $\alpha = 10.0$ & $\bm{108.13}$ & $27.21$ & $232.73$ & $0.6909$ \\
      \specialrule{1.5pt}{0pt}{0pt}
    \end{tabular}
  \end{table}
  
  \begin{figure}[htbp]
     \centering
     
     \begin{minipage}[b]{0.3\linewidth}
       \centering
      
       \includegraphics[scale=0.35]{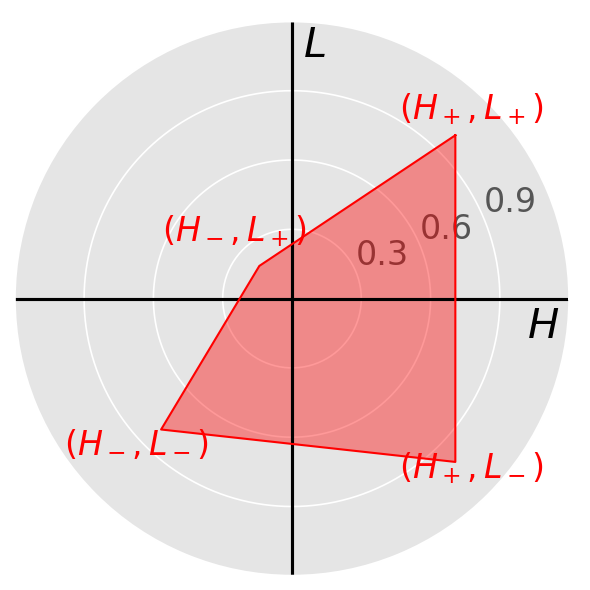}
       \caption*{$\alpha = 0.01$}
     \end{minipage}
     \begin{minipage}[b]{0.3\linewidth}
       \centering
      
       \includegraphics[scale=0.35]{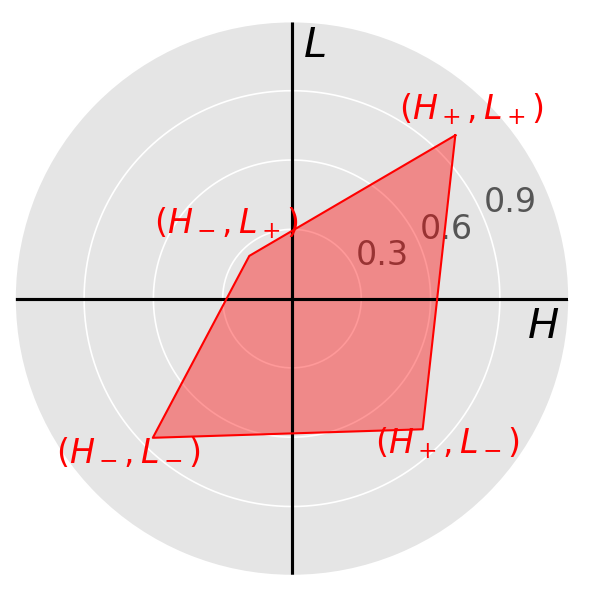}
       \caption*{$\alpha = 0.40$}
     \end{minipage}
     \begin{minipage}[b]{0.3\linewidth}
       \centering
    
       \includegraphics[scale=0.35]{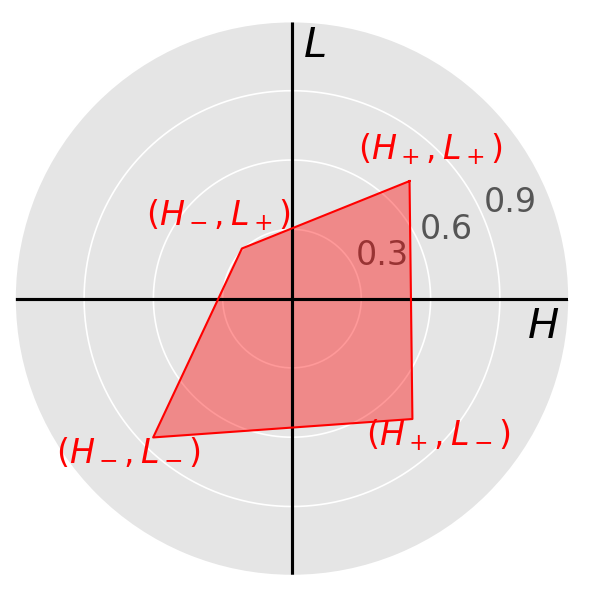}
       \caption*{$\alpha = 10.0$}
     \end{minipage}
     \caption*{(a) radar plot of $\overline{p}$.}

     \textsf{\small\bfseries Legend:}
     \raisebox{-5pt}{\includegraphics[scale=0.25]{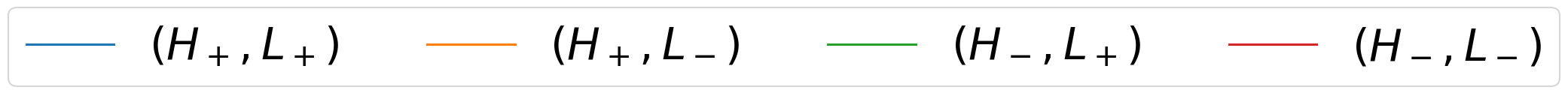}}
     
     \begin{minipage}[b]{0.3\linewidth}
       \centering
       \includegraphics[scale=0.2]{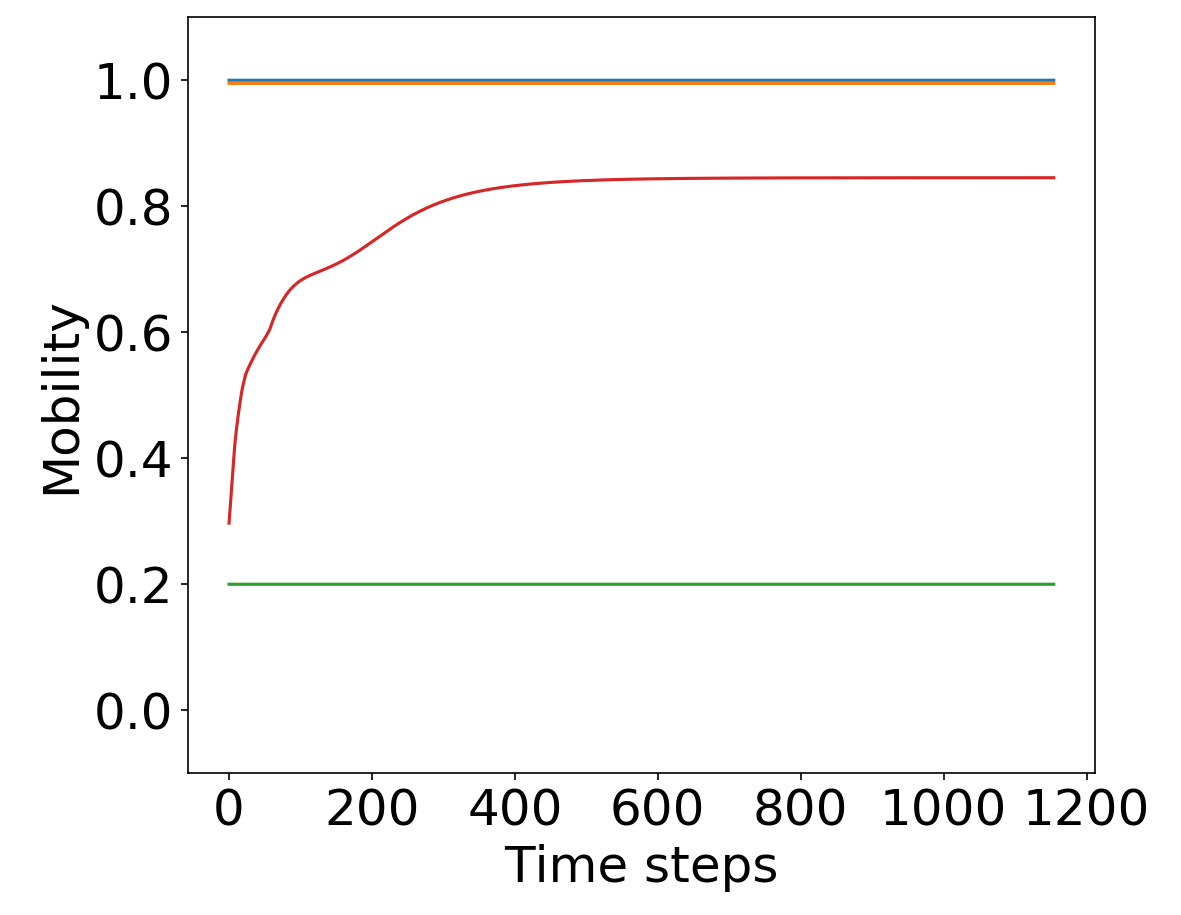}
       \caption*{$\alpha = 0.01$}
     \end{minipage}
     \begin{minipage}[b]{0.3\linewidth}
       \centering
       \includegraphics[scale=0.2]{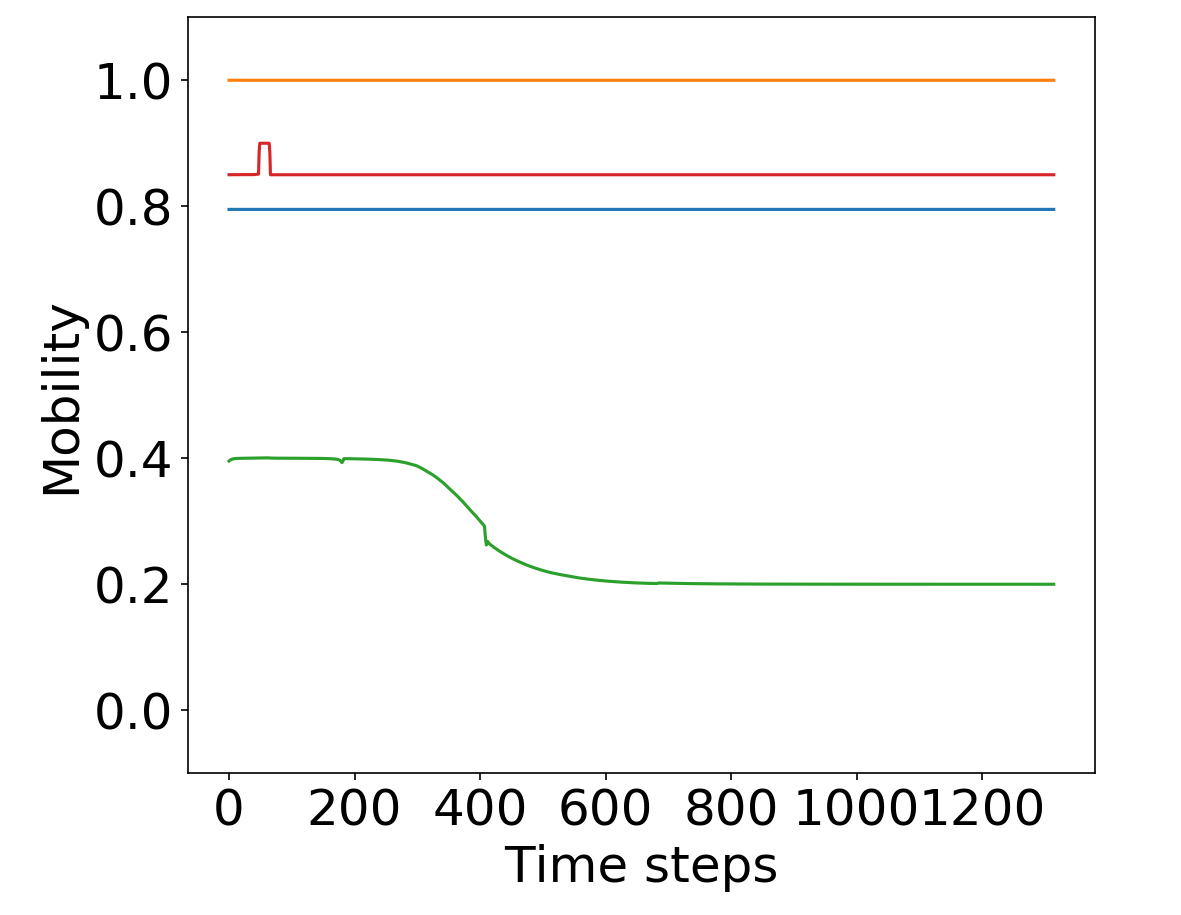}
       \caption*{$\alpha = 0.40$}
     \end{minipage}
     \begin{minipage}[b]{0.3\linewidth}
       \centering
       \includegraphics[scale=0.2]{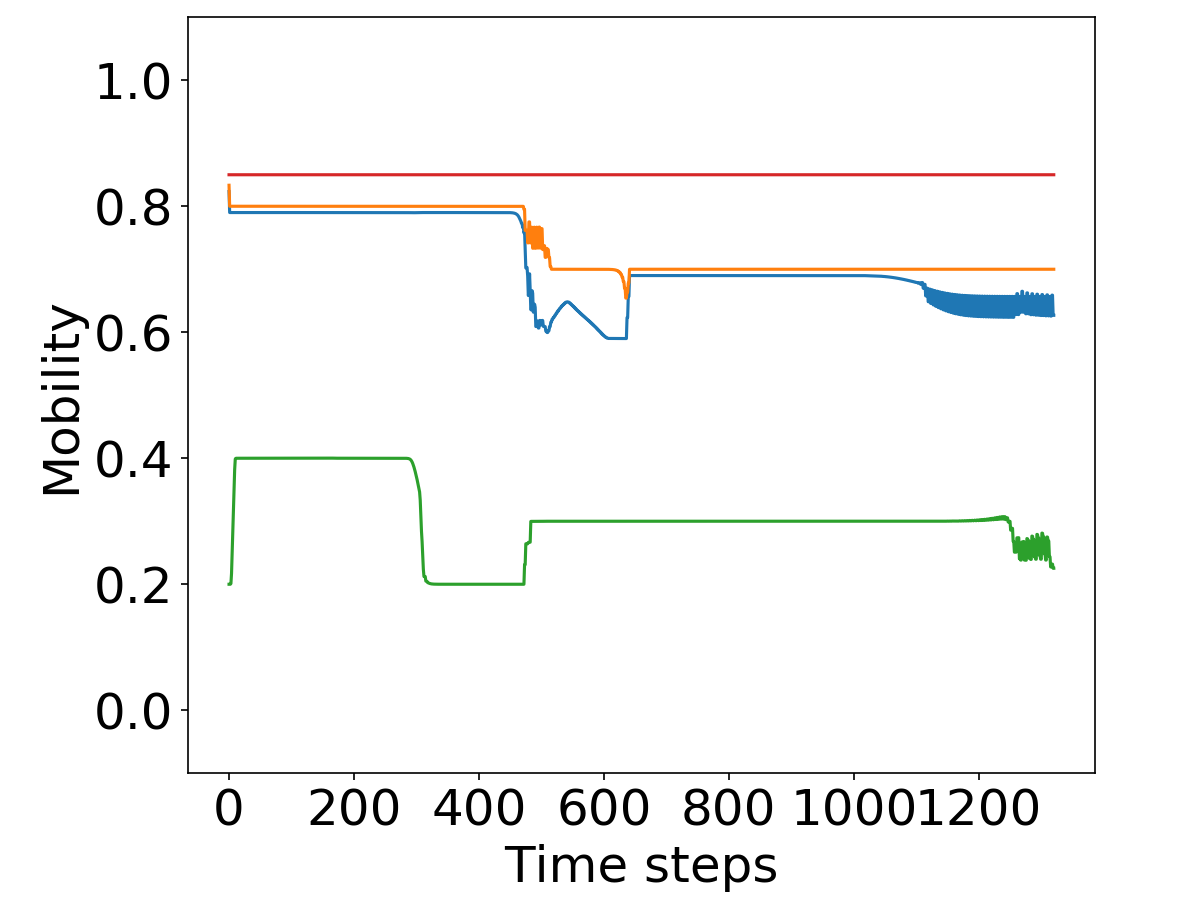}
       \caption*{$\alpha = 10.0$}
     \end{minipage}
     \caption*{(b) temporal change of $\overline{p}$.}
     
     \caption{Type-wise analysis of mean action $\overline{p}$ for different mixing ratios.}\label{fig:toy-outbreak-p}
  \end{figure}
 
\begin{figure}[htbp]
     \centering
     \textsf{\small\bfseries Legend:}
     \raisebox{-5pt}{\includegraphics[scale=0.25]{pics/legend.png}}
     
    
    \begin{minipage}[b]{0.48\linewidth}
      \centering
      \includegraphics[scale=0.3]{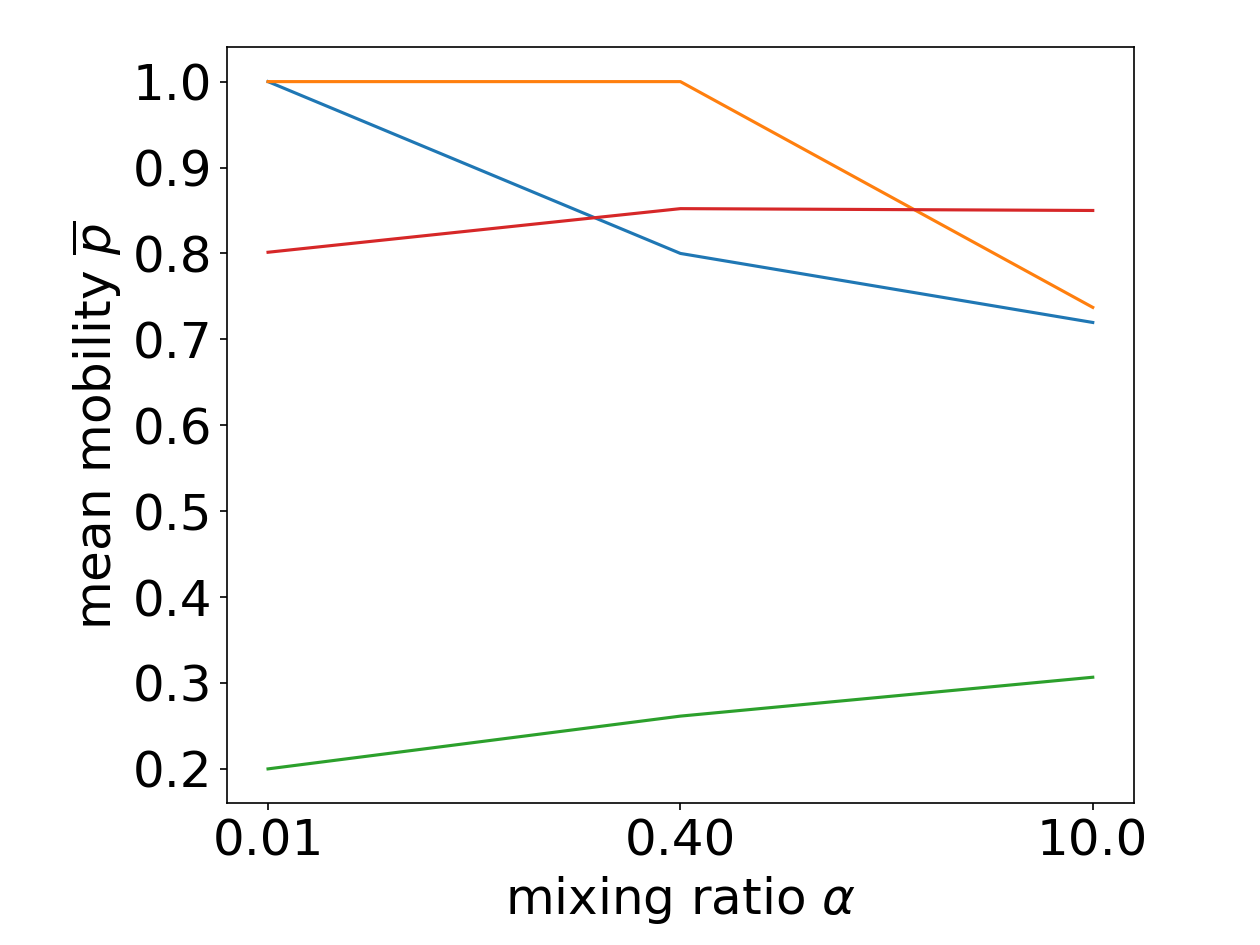}
    \end{minipage}
    \begin{minipage}[b]{0.48\linewidth}
      \centering
      \includegraphics[scale=0.3]{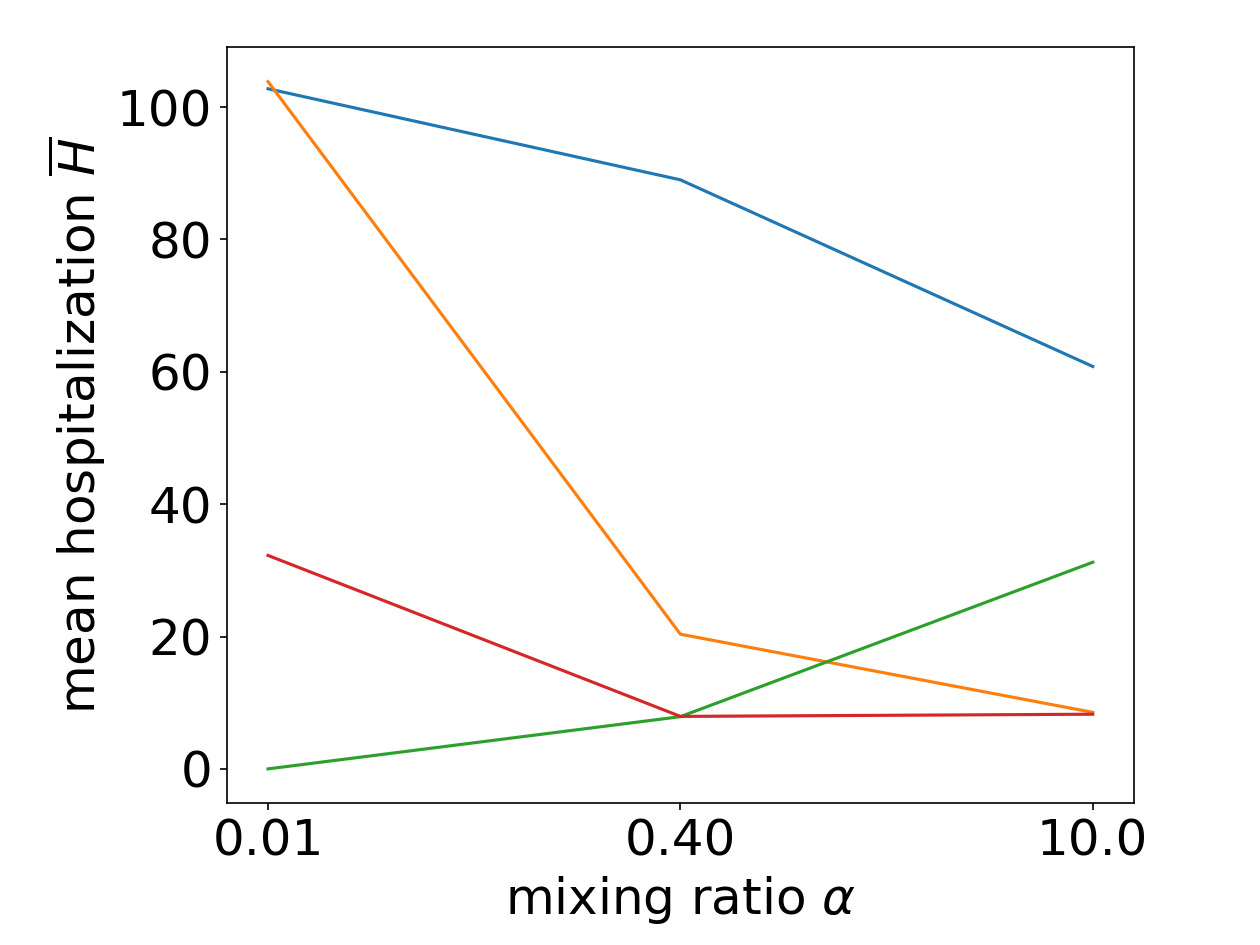}
    \end{minipage}
    \caption{Type-wise trend of change for $\overline{p}$ and $\overline{H}$ when mixing ratio increases.}\label{fig:toy-outbreak-change}
  \end{figure}

Despite the simplicity of the setting, this environment clearly illustrates different behavior patterns of agents of different regional types. For agents trained under different mixing ratios, the type-wise analysis of mean actions and mean hospitalization rates are shown in Figure \ref{fig:toy-outbreak-p} and \ref{fig:toy-outbreak-change}. It is observed that, when agents make decisions based largely on local rewards (i.e., in the case of $\alpha = 0.01$), they tend to select actions that best serve their own interests --- the $(H_-, L_+)$-region is extremely vulnerable to outbursts of infections, so its optimal policy is to enforce strict lockdown policies; the $(H_+, L_-)$-region cannot afford long-term lockdowns, so it prefers to largely open up inward traffic; the $(H_-, L_-)$-region suffers from both a pandemic outbreak and a large cumulative mobility loss, so it strictly blocks all inward traffic initially, but is forced to open up later to avoid cumulative lockdown penalty. This is clearly shown in the radar plot (\autoref{fig:toy-outbreak-p}(a)), where the colored area (representing mobility) leans towards $(H_+, L_-)$-direction, indicating that regions that can afford more patients but not strict lockdowns are more likely to open up.

The above behavioral patterns are similar to what we have observed in the very early stage of COVID-19 outbreak, where regional governments have not established unified collaboration protocols. The $(H_-, L_+)$-region corresponds to those regions in reality that have limited medical resources and relatively self-sufficient economies, so that they are willing to avoid local panic for medical resources at the cost of economic losses. The $(H_+, L_-)$-region represents those regions with export-oriented economies that rely heavily on inter-regional population and commodity mobility, so they will  not voluntarily enforce lockdown policies, but would rather maintain economic vitality even at the cost of increasing number of infections. The $(H_-, L_-)$-region, on the other hand, is a representative of under-developed regions that are the most vulnerable to a pandemic outbreak --- without help from other regions, they will soon fall into the dilemma that neither open-up nor lockdown is favorable. 

However, as the global reward gains more weight in agents' optimization objectives, their behavior gradually changes. The agents with higher lockdown-tolerance level ($L_+$) tend to share the lockdown burden with those agents who have lower lockdown-tolerance level ($L_-$), as they enforce stricter lockdown from the source region. Such changes will not be observed if each agent focuses on its own interest only, since for those $H_+$-regions that voluntarily share the burden, slightly larger infection numbers do not hurt them as badly as the extra lockdown costs; however, if they choose not to cooperate, the most vulnerable $(H_-, L_-)$-region has to block more traffic from $H_+$-regions to avoid imported infections, which vastly increases the lockdown cost of the $(H_-, L_-)$-region or even leads to oscillations in action. Therefore, enforcing extra lockdown by those $H_+$-regions against the source city helps more vulnerable cities to suffer less from pandemic outbreaks and/or cumulative mobility losses, which is indeed a kind of altruistic collaboration that promotes overall welfare and realizes better inter-regional pandemic control. The trend of changing behavior is illustrated in \autoref{fig:toy-outbreak-change}, and is also clear from the radar plot (\autoref{fig:toy-outbreak-p}(a)), as the colored area becomes more balanced among different types of regions.

The influence of the collaborative behavior of our IRC agents can also be reflected by the mean hospitalization matrix, as illustrated in \autoref{fig:toy-outbreak-change}. It is observed that, as the mixing ratio $\alpha$ increases, the number of hospitalized people in the $(H_-, L_-)$-region gradually decreases even when its mobility increases at the same time. Meanwhile, the majority of pandemic penalties are still undertaken by $H_+$-regions (since they still allow most inward mobility), which have more medical resources and are thus more resilient to such mild increases in infections. Therefore, our IRC agents indeed learn to balance the demand of regions with different regional types, and such balanced collaboration behavior is helpful to promote overall welfare (indicated by increased $\overline{R}_{\text{g}}$ and lower $\overline{H}$).

As a short summary, our IRC models successfully learn different policies for different mixing ratios, and the learned behavior shows an interpretable changing pattern of collaboration behavior --- regions with abundant medical resources and high resilience for pandemic outbreak will voluntarily enforce stricter lockdown policies to help those vulnerable regions at higher collaboration levels. It is also worth mentioning that, as the mixing ratio $\alpha$ increases, our IRC models are more likely to find policies that yield higher global rewards, which demonstrates the benefit of collaboration, and confirms the mixing ratio to be a good adjustable parameter representing the collaboration level.

\section{Conclusions and Future Work}\label{sec:conclusion}
In this paper, we analyze collaborative patterns of multi-region pandemic control behavior under different collaboration levels, using the optimal policies found by our specially-designed MARL method. We introduce a reward mixing ratio to abstract the inter-regional collaboration level that can be adjusted to reveal agents' different behavior difference under different collaboration levels. Experimental results in an exemplary environment show that, as the global reward accounts for a larger portion of each agent's objective, the actions and rewards of regions of different types become more balanced, and agents tend to learn policies that not merely promote their own interests, but also collaborate to help reduce the loss of other regions. One major take-away message here is that higher collaboration level will lead to better and more balanced pandemic control policies, in that regions will voluntarily help to coordinate the demands of different regions, and regions more resilient to lockdown tend to sacrifice a mild portion of its own utility to protect more vulnerable regions.

Our framework provides a novel computational perspective for understanding and promoting collaboration for the supply of public goods. As a representative type of them, pandemic control cannot be supplied with a few regions or through a few isolated policies, while different attitudes towards inter-regional collaboration enables different pandemic control policies, and will possibly lead to different outcomes. This reminds us that, to better cope with global catastrophes like COVID-19 and more efficiently provide public goods to promote global welfare, international collaboration should be encouraged and all related parties should be called for to actively assume international responsibilities, so that the strong will protect the weak by sharing the losses. The legitimacy of such collaborative policies that restrict individual liberty can be justified by the optimality guarantee of the computational methods. We believe this framework can be generalized to analyze the supply of other public goods that involves collaboration.

The model of this paper is limited in that it regards inter-regional mobility control as the only form of action. However,  regions' within-regional pandemic control strategies will also induce epidemiological and economic costs, which also potentially influences inter-region collaboration. For example, a destination region may be more willing to accept inward mobility if the source region enforces better control within itself. Therefore, a more comprehensive modelling is left as future work. 

\bibliographystyle{unsrtnat}
\bibliography{main}

\newpage
\appendix
\section{Environmental Modelling}
\subsection{Mobility}\label{appendix:mobility}
At each time step $t$, the mobility demands of all $n$ regions are represented by a matrix $M^t_\text{d} \in \mathbb{R}^{n \times n}$, where $M^t_{\text{d},i,j}$ denotes the mobility demand from region $i$ to region $j$.

All actions at time step $t$ are arranged as a matrix $p^t \in [0,1]^{n \times n}$, where $p^t_{i,j}$ denotes the proportion of mobility demand that can be fulfilled from region $i$ to region $j$, which is decided by the destination region $j$. Therefore, region $j$ is allowed to determine the $j$\textsuperscript{th} column $p_j \in [0,1]^n$ in matrix $p$.

Eventually, the actual allowed mobility $M^t_{\text{a}}$ at time step $t$ is calculated by $M^t_{\text{a},i,j} = M^t_{\text{d},i,j} p^t_{i,j}$.

\subsection{Pandemic Transmission}\label{appendix:transmission}
The pandemic transmission model is based on the Susceptible-Infected-Hospitalized-Recovered (SIHR) model proposed in \cite{DURLECA}. The \textit{pandemic state} of region $i$ at time step $t$ is $E^t_i= (S^t_i, I_i^t, H_i^t, R_i^t)$, which consists of the number of susceptible, infected, hospitalized, and recovered population within that region, respectively. Agents are able to observe the number of hospitalization and recovery, but cannot distinguish between susceptible (healthy) people and infected people that are both asymptomatic. Therefore, the \textit{visible pandemic state} of region $i$ at time step $t$ is $E^t_{\text{v},i}= (S^t_i+I_i^t, H_i^t, R_i^t)$.

At each step $t$, pandemic transmission is calculated in two phases: mobility happens first, and then the pandemic spreads within each region.

At the mobility happening stage, people move between regions according to actual allowed mobility $M^t_\text{a}$, calculated in the previous section. After the mobility happening stage, the intermediate state becomes $\hat{E}^t_i = (\hat{S}^{t}_i,\hat{I}^{t}_i,\hat{R}_i^t)$, which is given by
\begin{align}
    E^t_{\text{s},i} = E^t_i - \sum_j \frac{M^t_{\text{a},i,j}}{N^t_i} E^t_i,\quad
    E^t_{\text{m},i} = \sum_j \frac{M^t_{\text{a},j,i}}{N^t_j} E^t_j,\quad
    \hat{E}^t_i = E^t_{\text{s},i} + E^t_{\text{m},i}.
\end{align}
Here $N_i^t$ is the population of region $i$ at time step $p$, $E^t_{\text{m},i}$ stands for the pandemic state of the moving people, and $E^t_{\text{s},i}$ stands for the pandemic state of the staying people.

At the pandemic spreading stage, the eventual pandemic state is calculated by
\begin{align}
    &S^{t+1}_i=\hat{S}^{t}_i-\frac{\beta^t_{\text{s},i}S^t_{\text{s},i}I^{t}_{\text{s},i}}{N^t_{\text{s},i}}-\frac{\beta^t_{\text{m},i}S^t_{\text{m},i}I^{t}_{\text{m},i}}{N^t_{\text{m},i}},\nonumber\\
    &I^{t+1}_i=\hat{I}^{t}_i+\frac{\beta^t_{\text{s},i}S^t_{\text{s},i}I^{t}_{\text{s},i}}{N^t_{\text{s},i}}+\frac{\beta^t_{\text{m},i}S^t_{\text{m},i}I^{t}_{\text{m},i}}{N^t_{\text{m},i}}-\gamma \hat{I}^{t}_i,\\
    &H^{t+1}_i=H^t_i+\gamma_i^t \hat{I}_i^t-\theta_i^t H_i^t,\nonumber\\
    &R_i^{t+1}=\hat{R}_i^t+\theta_i^t H_i^t,\nonumber
\end{align}
where $\beta^t_{i}$ is the transmission rate of the susceptible people, $\gamma^t_i$ is the hospitalization rate of the infected people, and $\theta^t_i$ is the recovery rate of the hospitalized people.

\subsection{Tolerance Levels}\label{appendix:tolerance-levels}
The local reward of an agent's action on that region is a combination of the pandemic-spread cost and the mobility-control cost. The pandemic-spread cost $C_{\text{p}, i}$ for agent $i$ is given by
\begin{equation}
  C_{\text{p},i} = k_\text{h} \exp \left( \frac{H_i}{H_{0,i}} \right),
\end{equation}
where $k_\text{h}$ is the hyperparameter that abstracts the cost induced by the first observed unit of infection within the region, while $H_{0,i}$ is a hyperparameter called \textit{pandemic-tolerance level} that determines the exponentially increasing rate of the cost as the number of hospitalized patients increases. The mobility-control cost $C_{\text{m},i}$ of region $i$ is given by
\begin{align}
    M^t_{\text{d},i} &= \sum_{j} M^t_{\text{d},j,i} \text{ }, \quad M^t_{\text{a},i}=\sum_{j} M^t_{\text{a},j,i} \text{ },\nonumber\\
    L^t_i &= \sum_{\tau=0}^{t-1}\lambda^{t-\tau}\frac{M_{\text{d},i}^{t-\tau}-M_{\text{a},i}^{t-\tau}}{\bar{M_{\text{d},i}}},\\
    C^t_{\text{m},i} &= \exp\left( \frac{L_i^t}{L_{0,i}} \right) \frac{M_{\text{d},i}^{t-\tau} - M_{\text{a},i}^{t-\tau}}{\bar{M_{\text{d},i}}}.\nonumber
\end{align}
Here $M^t_{\text{d},i}$ is the total mobility demand into region $i$, while $M^t_{\text{a},i}$ is the actual amount of mobility that is allowed to enter region $i$ at time step $t$; $L^t_{0,i}$ is the cumulative penalty caused by continuous mobility restriction of the same region, where $\lambda$ is the temporal discount factor of historical penalties; $L_0$ is the hyperparameter called \textit{lockdown-tolerance level} that determines the exponentially increasing rate of the cost as the allowed amount of mobility decreases.

\section{Detailed Experimental Results}
  Detailed experimental results are displayed below in \autoref{fig:toy-outbreak-table}.
  
  \begin{table}[H]
    \centering
    \caption{Type-wise behavior patterns for agents of different types.}\label{fig:toy-outbreak-table}
    
    \vspace*{8pt}
    \caption*{(a) Type-wise comprison table of mean action $\overline{p}$ for different mixing ratios.}
    \begin{minipage}[b]{0.3\linewidth}
      \centering
      \begin{tabular}{c|c|c}
         \specialrule{1.5pt}{0pt}{0pt}
         $\overline{\bm{p}}$ & $\bm{L_+}$ & $\bm{L_-}$ \\
         \specialrule{0.4pt}{0pt}{0pt}
         $\bm{H_+}$ & $1.000$ & $1.000$ \\
         \specialrule{0.4pt}{0pt}{0pt}
         $\bm{H_-}$ & $0.200$ & $0.801$ \\
         \specialrule{1.5pt}{0pt}{0pt}
      \end{tabular}
      
      \caption*{$\alpha = 0.01$}
    \end{minipage}
    \begin{minipage}[b]{0.3\linewidth}
      \centering
      \begin{tabular}{c|c|c}
         \specialrule{1.5pt}{0pt}{0pt}
         $\overline{\bm{p}}$ & $\bm{L_+}$ & $\bm{L_-}$ \\
         \specialrule{0.4pt}{0pt}{0pt}
         $\bm{H_+}$ & $0.800$ & $1.000$ \\
         \specialrule{0.4pt}{0pt}{0pt}
         $\bm{H_-}$ & $0.261$ & $0.852$ \\
         \specialrule{1.5pt}{0pt}{0pt}
      \end{tabular}
      
      \caption*{$\alpha = 0.40$}
    \end{minipage}
    \begin{minipage}[b]{0.3\linewidth}
      \centering
      \begin{tabular}{c|c|c}
         \specialrule{1.5pt}{0pt}{0pt}
         $\overline{\bm{p}}$ & $\bm{L_+}$ & $\bm{L_-}$ \\
         \specialrule{0.4pt}{0pt}{0pt}
         $\bm{H_+}$ & $0.720$ & $0.737$ \\
         \specialrule{0.4pt}{0pt}{0pt}
         $\bm{H_-}$ & $0.307$ & $0.850$ \\
         \specialrule{1.5pt}{0pt}{0pt}
      \end{tabular}
      
      \caption*{$\alpha = 10.0$}
    \end{minipage}
    
    \vspace*{5pt}
    \caption*{(b) Type-wise comprison table of mean hospitalization $\overline{H}$ for different mixing ratios.}
    \begin{minipage}[b]{0.3\linewidth}
      \centering
      \begin{tabular}{c|c|c}
         \specialrule{1.5pt}{0pt}{0pt}
         $\overline{\bm{H}}$ & $\bm{L_+}$ & $\bm{L_-}$ \\
         \specialrule{0.4pt}{0pt}{0pt}
         $\bm{H_+}$ & $102.76$ & $103.84$ \\
         \specialrule{0.4pt}{0pt}{0pt}
         $\bm{H_-}$ & $0.00$ & $32.25$ \\
         \specialrule{1.5pt}{0pt}{0pt}
      \end{tabular}
      
      \caption*{$\alpha = 0.01$}
    \end{minipage}
    \begin{minipage}[b]{0.3\linewidth}
      \centering
      \begin{tabular}{c|c|c}
         \specialrule{1.5pt}{0pt}{0pt}
         $\overline{\bm{H}}$ & $\bm{L_+}$ & $\bm{L_-}$ \\
         \specialrule{0.4pt}{0pt}{0pt}
         $\bm{H_+}$ & $89.00$ & $20.35$ \\
         \specialrule{0.4pt}{0pt}{0pt}
         $\bm{H_-}$ & $7.89$ & $7.93$ \\
         \specialrule{1.5pt}{0pt}{0pt}
      \end{tabular}
      
      \caption*{$\alpha = 0.40$}
    \end{minipage}
    \begin{minipage}[b]{0.3\linewidth}
      \centering
      \begin{tabular}{c|c|c}
         \specialrule{1.5pt}{0pt}{0pt}
         $\overline{\bm{H}}$ & $\bm{L_+}$ & $\bm{L_-}$ \\
         \specialrule{0.4pt}{0pt}{0pt}
         $\bm{H_+}$ & $60.80$ & $8.54$ \\
         \specialrule{0.4pt}{0pt}{0pt}
         $\bm{H_-}$ & $31.22$ & $8.26$ \\
         \specialrule{1.5pt}{0pt}{0pt}
      \end{tabular}
      
      \caption*{$\alpha = 10.0$}
    \end{minipage}
  \end{table}

\end{document}